\documentclass{article}

\usepackage{PRIMEarxiv}

\usepackage[utf8]{inputenc} 
\usepackage[T1]{fontenc}    
\usepackage{hyperref}       
\usepackage{url}            
\usepackage{booktabs}       
\usepackage{amsfonts}       
\usepackage{nicefrac}       
\usepackage{microtype}      
\usepackage{lipsum}
\usepackage{fancyhdr}       
\usepackage{graphicx}       
\graphicspath{{media/}}     

\usepackage{caption}
\usepackage{subcaption}
\usepackage{cuted}
\usepackage{amsmath,amsfonts}
\usepackage[ruled,vlined]{algorithm2e}

\usepackage{enumitem}

\usepackage[ruled,vlined]{algorithm2e}
\usepackage{booktabs}
\DeclareMathAlphabet\mathbfcal{OMS}{cmsy}{b}{n}
\usepackage{bm}
\usepackage{authblk}

\usepackage[table,xcdraw]{xcolor}
\usepackage{makecell}
\usepackage[normalem]{ulem}
\useunder{\uline}{\ul}{}

\title{Capsules as viewpoint learners for human pose estimation
}

\author{
  Nicola Garau, Nicola Conci \\
  University of Trento \\
  CNIT \\
  \texttt{nicola.garau, nicola.conci@unitn.it}
}

\begin{document}
\maketitle

\vspace*{-20pt}

\begin{abstract}
The task of human pose estimation (HPE) deals with the ill-posed problem of estimating the 3D position of human joints directly from images and videos. In recent literature, most of the works tackle the problem mostly by using convolutional neural networks (CNNs), which are capable of achieving state-of-the-art results in most datasets. We show how most neural networks are not able to generalize well when the camera is subject to significant viewpoint changes. This behaviour emerges because CNNs lack the capability of modelling viewpoint equivariance, while they rather rely on viewpoint invariance, resulting in high data dependency. Recently, capsule networks (CapsNets) have been proposed in the multi-class classification field as a solution to the viewpoint equivariance issue, reducing both the size and complexity of both the training datasets and the network itself. In this work, we show how capsule networks can be adopted to achieve viewpoint equivariance in human pose estimation. We propose a novel end-to-end viewpoint-equivariant capsule autoencoder that employs a fast Variational Bayes routing and matrix capsules. We achieve state-of-the-art results for multiple tasks and datasets while retaining other desirable properties, such as greater generalization capabilities when changing viewpoints, lower data dependency and fast inference. Additionally, by modelling each joint as a capsule, the hierarchical and geometrical structure of the overall pose is retained in the feature space, independently from the viewpoint. We further test our network on multiple datasets, both in the RGB and depth domain, from seen and unseen viewpoints and in the viewpoint transfer task. This work is an extended version of DECA \cite{garau2021deca}. The original paper and code can be found at the link below \footnote{https://paperswithcode.com/paper/deca-deep-viewpoint-equivariant-human-pose}.
\end{abstract}

\begin{figure}
    \centering
    \includegraphics[width=\textwidth]{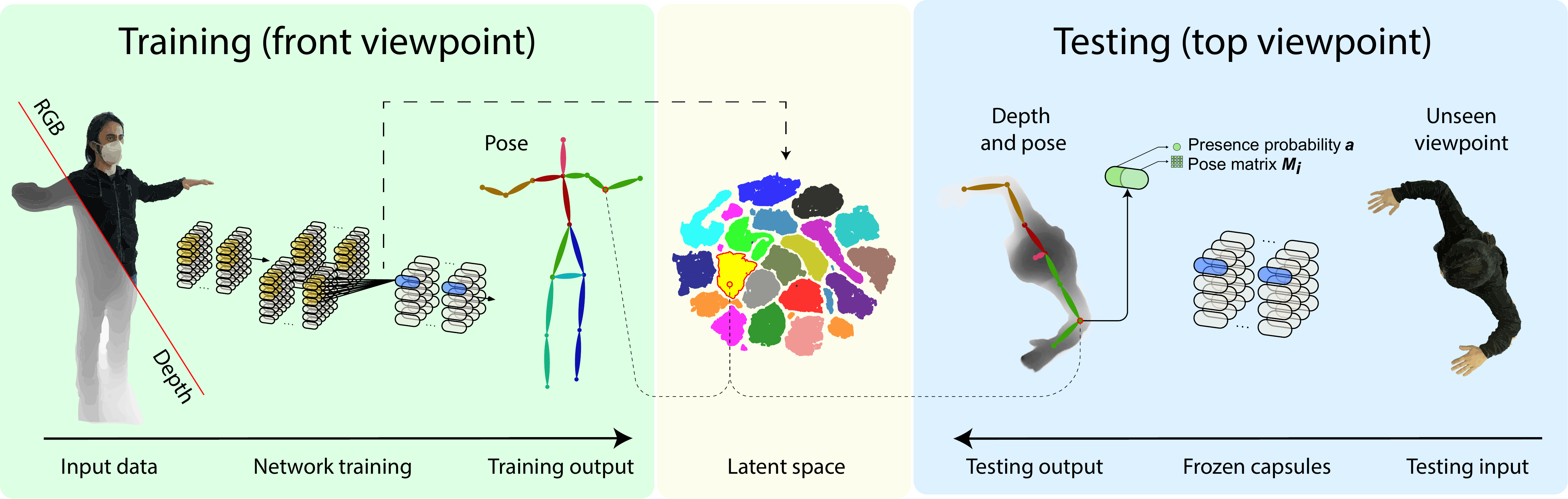}
    \captionof{figure}{\textbf{[Better seen in color].} General description of the presented method. In green (left) we show the training procedure; input data (RGB or depth map images) from a single viewpoint (e.g. front) are fed to a capsule autoencoder, which learns to reconstruct the 3D pose in the output layer. Additionally, we constrain every capsule in the last layer to focus on a single joint, simultaneously modelling its presence probability $a$ and a pose matrix $M_i$. At testing time (light blue, right), we feed the trained capsules an image belonging from a completely different and unseen viewpoint (e.g. top). The proposed network is able to correctly guess the pose, since it has also learned the concept of viewpoint, embedded in each capsule. Results show good and coherent separation in the latent space (light yellow, center) for each viewpoint, and state of the art results on multiple datasets and for the viewpoint-transfer task.}
    \label{fig:teaser_new}
\end{figure}


\section{Introduction}
\label{sec:introduction}

\begin{figure}
    \centering
    \includegraphics[width=\textwidth]{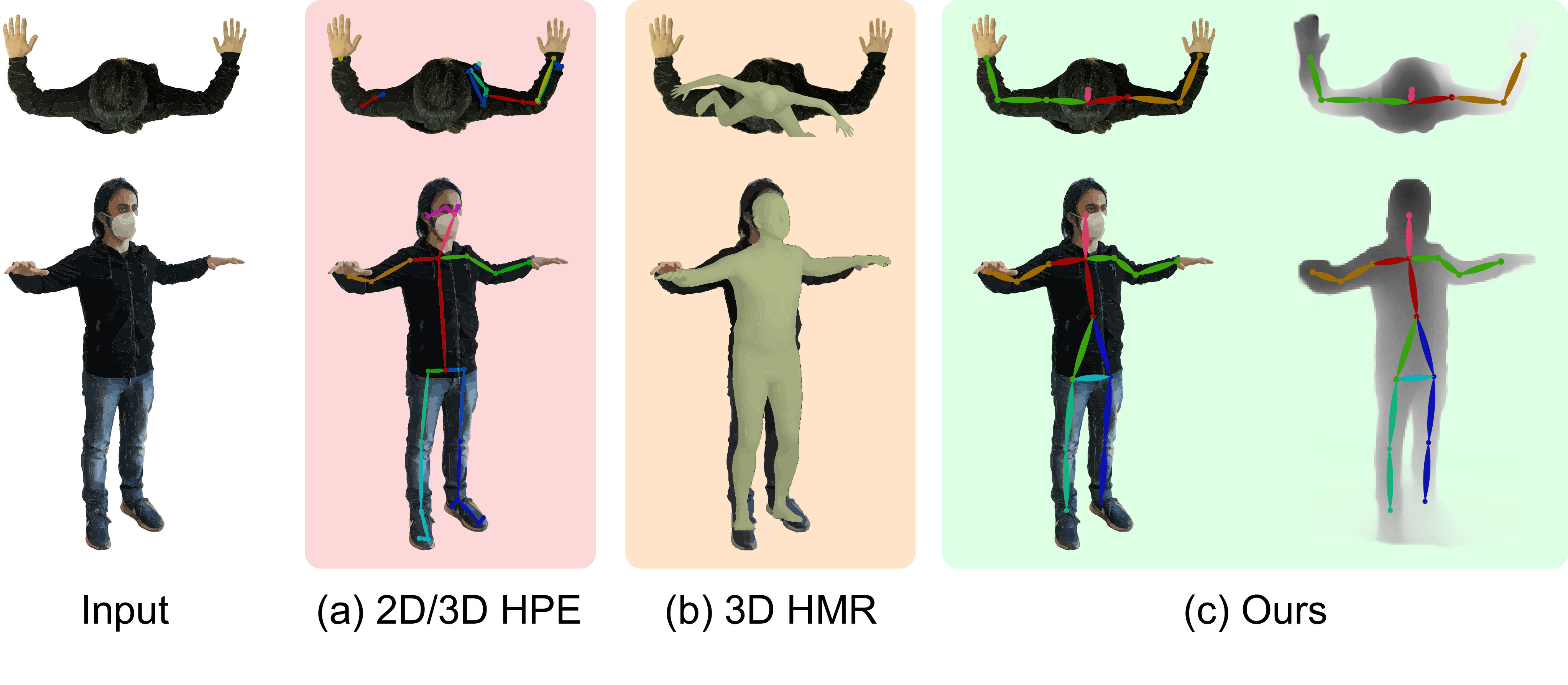}
    \captionof{figure}{\textbf{[Better seen in color].} Overview of the proposed solution. Two different views of the same subject are shown for each image: (a) 2D/3D Human Pose Estimation (HPE) and (b) 3D Human Mesh Recovery (HMR) methods achieve good accuracy on the front-view  (second row). Changing the viewpoint turns into performance degradation (first row). Our method (c) promotes viewpoint equivariance, showing good results in both the RGB and depth domains.}
    \label{fig:teaser}
\end{figure}

Human pose estimation is key for many applications, such as action recognition, animation, gaming, to name a few \cite{kalfaoglu2020late, starke2019neural, shotton2011real}. 
State of the art methods \cite{cao2017realtime,tome2017lifting} that rely on RGB images can correctly localize human joints (e.g. torso, elbows, knees) in images, also in presence of occlusions. However, they tend to fail when dealing with challenging scenarios. The top-view perspective, in particular, turns out to be a difficult task; on the one hand, it causes the largest amount of joints occlusions, and on the other hand, it suffers the scarcity of suitable training data, as shown in Fig. \ref{fig:teaser}. 

When presented with unseen viewpoints, humans display a remarkable ability to estimate human poses, even in the presence of occlusions and unconventional joints configurations. This is not always true in computer vision. In fact, available methods are trained in relatively constrained settings \cite{Joo_2017_TPAMI}, with a limited variability between different viewpoints. Limited data, especially from the top-viewpoint, along with limited capabilities of modeling the hierarchical and geometrical structure of the human pose, results in poor generalization capabilities.

This generalization problem, known as the \textit{viewpoint problem}, depends on how the network activations vary with the change of the viewpoint, usually after a transformation (translation, scaling, rotation, shearing). Convolutional Neural Networks (CNNs) scalar activations are not suitable to effectively manage these viewpoint transformations, thus needing to rely on max-pooling and aggressive data augmentation \cite{cohen2018spherical, haque2016towards,moon2018v2v,xiong2019a2j}. By doing so, CNNs aim at achieving \textit{viewpoint invariance}, defined as 
\begin{alignat}{3}
  f(Tx) &= f(x) \label{eq:invariance}
\end{alignat}

According to this formulation, applying a viewpoint transformation \textit{T} on the input image $x$, \textit{does not change} the outcome of the network activations.

However, a more desirable property would be to capture and retain the transformation \textit{T} applied to the input image $x$, thus obtaining a network that is aware of the different transformations applied to the input. Being able to model network activations that change in a structured way according to the input viewpoint transformations is also called \textit{viewpoint equivariance} and it is defined as:

\begin{alignat}{3}
  f(Tx) &= Tf(x). \label{eq:equivariance} 
\end{alignat}

This is achieved by introducing \textit{capsules}: groups of neurons that explicitly encode the intrinsic viewpoint-invariant relationship existing between different parts of the same object. 
Capsule networks (CapsNets) can learn part-whole relationships between so-called \textit{entities} across different viewpoints \cite{hinton2011transforming, sabour2017dynamic, hinton2018matrix}, similarly to how our visual cortex system operates, according to the recognition-by-components theory \cite{biederman1987recognition}.
Unlike traditional CNNs, which usually retain viewpoint invariance, capsule networks can explicitly model and jointly preserve a viewpoint transformation \textit{T} through the network activations, achieving \textit{viewpoint equivariance} (Eq. \ref{eq:equivariance}). 

Developing viewpoint-equivariant methods for 3D HPE networks leads to multiple advantages: (i) the learned model is more robust, interpretable, and suitable for real-world applications, (ii) the viewpoint is treated as a learnable parameter, allowing to disentangle the 3D data of the skeleton from each specific view, (iii) the same annotated data can be used to train a network for different viewpoints, thus less training data is required.

In this work, we address the problem of viewpoint-equivariant human pose estimation from single depth or RGB images. A more comprehensive overview of the presented solution can be seen in Fig. \ref{fig:teaser_new}. Our contribution is summarised as follows:
\begin{itemize}
    \item We present \textbf{a novel Deep viewpoint-Equivariant Capsule Autoencoder architecture (DECA)} which jointly addresses multiple tasks, such as 3D and 2D human pose estimation.

    \item We show how our network works with limited training data, no data augmentation, and across different input domains (RGB and depth images).

    \item We show how the feature space organization, defined by routing the input information to build capsule entities, improves when the tasks are jointly addressed.

    \item We evaluate our method on the ITOP \cite{haque2016towards} dataset for the depth domain and on the PanopTOP31K \cite{garau2021panoptop} dataset for the RGB domain. We establish a new baseline for the viewpoint transfer task and in the RGB domain.
\end{itemize}



\section{Related work}
\label{sec:related}

In recent years, human pose estimation has been a subject of multiple studies, particularly for real-time 2D HPE \cite{cao2017realtime}, 3D HPE \cite{tome2017lifting} and human mesh recovery (HMR) approaches \cite{kolotouros2019learning, kocabas2019vibe}.
In this work, we focus on HPE from single views, using either RGB \cite{cao2017realtime,he2017mask} or depth images \cite{haque2016towards,moon2018v2v,xiong2019a2j}.

\textit{\textbf{Viewpoint-invariant HPE from RGB images.}} 3D HPE usually leverages on additional cues, such as 2D predictions \cite{tome2017lifting, wang20203d, tekin2017learning}, multiple images \cite{zhou2016sparseness}, pre-trained models \cite{katircioglu2018learning} and pose dictionaries \cite{sanzari2016bayesian}. Other recent works aim at end-to-end, learning-based 3D HPE \cite{rogez2019lcr, tian2019densely, liu2020feature}.
 In the RGB domain, common HPE datasets such as Human3.6M \cite{ionescu2014human}, provide images from multiple views, like front-view or side-view, while the top-view component is generally missing. 
It is then evident that the lack of suitable multi-view (top-view in particular) data implies that 
state-of-the-art methods \cite{cao2017realtime,tome2017lifting, kolotouros2019learning,kocabas2019vibe} necessarily perform poorly when presented with an unseen viewpoint at test time, as shown in Fig. \ref{fig:teaser}(a).

\textit{\textbf{Viewpoint-invariant HPE from depth images.}}
Viewpoint invariant HPE methods have been developed using depth images \cite{haque2016towards,moon2018v2v,xiong2019a2j} from top-view and side-view, using datasets like the K2HPD Body Pose Dataset \cite{wang2016human} and the ITOP dataset \cite{haque2016towards}.
To take advantage of the 3D information encoded in 2D depth images, one recent research trend is to resort to 3D deep learning. The paid efforts can be generally categorized into 3D CNN-based and point-set-based families.
To enhance the 3D proprieties of depth data and compute more significant features, current methods rely on 3D CNNs \cite{haque2016towards,moon2018v2v} or 2D CNNs with dense features \cite{xiong2019a2j}.

3D CNN-based methods \cite{haque2016towards,moon2018v2v} perform a voxelization operation on pixels to transform them into 3D objects. To process the 3D data, each network performs costly 3D convolutions on the input data. These operations are responsible for the high computational burden and the difficulty to properly tune a high number of parameters in 3D CNNs.
In the domain of 2D CNNs, Xiong et al.\cite{xiong2019a2j} capture the 3D structure by computing dense features in an ensemble way, thus avoiding computationally intensive CNN layers, but they still rely on a backbone pre-trained network to extract 2D features.
Still, the above-mentioned approaches usually achieve weak viewpoint-invariance but fail to model viewpoint-equivariance. 
Moreover, we argue that the 3D geometry of the data should be interpreted by the network without relying on the voxelization embedding, or a 2D pre-trained feature extraction network. 

\textit{\textbf{Capsule networks for HPE.}} Capsule networks have shown the ability to model the geometric nature of training data thanks to the network structure and features \cite{sabour2017dynamic, hinton2018matrix, kosiorek2019stacked}.
Sabour et al.., introduce a routing algorithm for vector capsules, called \textit{routing-by-agreement} as a better max-pooling substitute.
Hinton et al. \cite{hinton2018matrix} further improve accuracy through a more complex matrix capsule structure and an Expectation-Maximization routing (\textit{EM-routing}) for capsules. Unfortunately, the EM-routing and the $4\times4$ pose matrix embedded in the capsule contribute to increasing the training time, when compared to both CNNs and vector CapsNets. 
Kosiorek et al. \cite{kosiorek2019stacked} introduce for the first time an unsupervised capsule-based autoencoder. Ribeiro et al. in \cite{ribeiro2020capsule} build upon the EM-routing version of capsule by proposing for the first time a Variational Bayes capsule routing (VB routing) fitting a mixture of transforming Gaussians. They present state-of-the-art results using $\sim50\%$ fewer capsules, achieving both performance gain and network complexity reduction. However, all the mentioned works only consider small datasets, such as MNIST, smallNORB, and CIFAR-10 for benchmarking. 

In the RGB domain, Ramírez \cite{ramirez2020bayesian} tackles the problem of RGB HPE using dynamic vector capsule networks \cite{sabour2017dynamic} to solve the 3D HPE problem in an end-to-end fashion. However, their work only exploits lateral viewpoints from the Human3.6M dataset and only considering RGB data.

In this work, we use matrix capsules \cite{hinton2018matrix}, along with a different capsule routing algorithm and a new encoding-decoding pipeline with GELU activations. We argue that matrix capsules are better suited than vector capsules for the 3D HPE task, as the $4\times 4$ pose matrix used for the routing can capture 3D geometry better than a dynamic vector structure.

\section{Method}
\label{sec:method}

\begin{figure*}[]
    \centering
    \includegraphics[width=\textwidth]{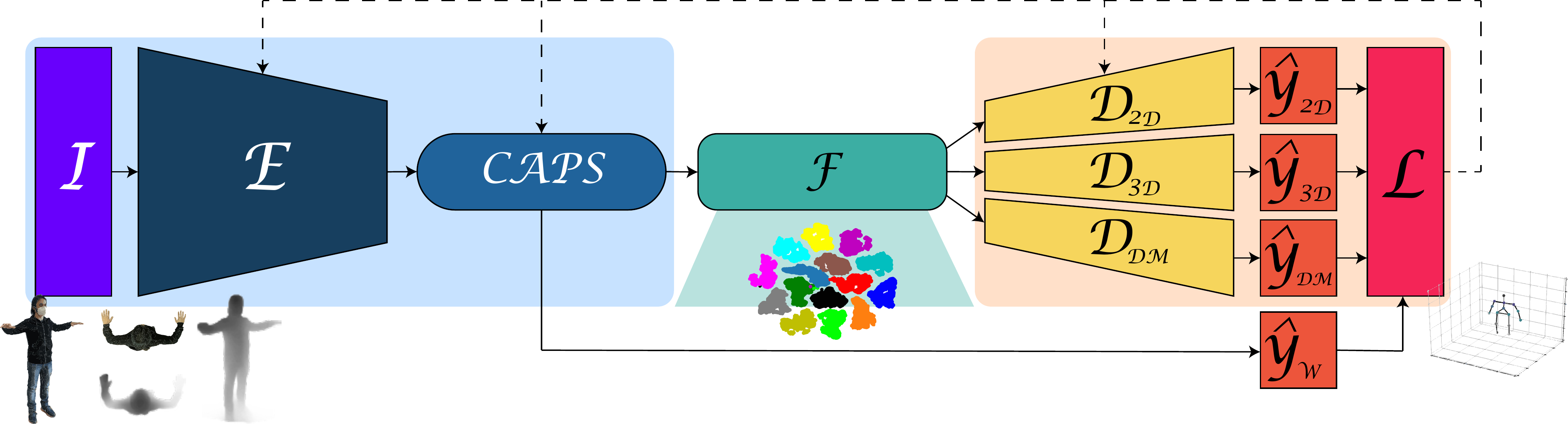}
    \captionof{figure}{\textbf{[Better seen in color].} Overview of the proposed architecture. In light blue, the encoding module (Input, CNN encoder, Capsule layers), in green the interpretable feature space with capsule entities, in light orange the decoding module (fully connected decoders with multiple tasks and self-balancing loss).}
    \label{fig:network}
\end{figure*}

We now analyze the proposed autoencoder, DECA, starting with the capsule encoder and the multi-task decoders.
DECA can be trained end-to-end, without any pre-training or data augmentation, and it works in real-time in the inference phase. An overview of the proposed architecture is shown in Fig. \ref{fig:network}.

\subsection{Capsule encoder}
\label{subsec:encoder}
The encoding module of the network (light blue in Fig. \ref{fig:network}) is divided in: (i) an input pre-processor $I$, (ii) a CNN encoder $E$ and (iii) four layers of Matrix Capsules with Variational Bayes Routing \cite{ribeiro2020capsule}.

 (i) $I$ is a  layer which normalizes the different type of data (RGB images, depth images, top-view, side-view, free-view) in the interval $[0,1]$. 

(ii) The normalised input is then forwarded to a CNN encoder $E$, built using four convolutional layers with inputs $[N_{ch}, 64, 128, 256]$, instance normalisation and GELU activations \cite{hendrycks2016gaussian}, as shown in Eq. \ref{eq:GELU}. $N_{ch}$ is the number of channels, which may vary depending on the input.

\begin{equation} \label{eq:GELU} 
    \begin{split}
    \text{GELU}(x) &\approx 0.5x(1 + \tanh{\Big[ \sqrt{\frac{2}{\pi}}(x + 0.044715x^3) \Big]} )
    \end{split}
\end{equation}

(iii) The output of the CNN encoder $E$ feeds our capsule layers.
 It has been shown in previous works \cite{sabour2017dynamic, hinton2018matrix, kosiorek2019stacked} that capsules provide a superior understanding of the viewpoint and the relationship between parts and parent objects, thus aiming at true viewpoint equivariance. Given the multiple degrees of freedom of each joint, we adopt the matrix capsules model \cite{hinton2018matrix} instead of vector capsules \cite{sabour2017dynamic}, enriching the description of single joints as hierarchically linked capsule entities. We deploy the novel capsule routing based on Variational Bayes (VB) \cite{ribeiro2020capsule}, which is proven to speed up the training of our matrix capsules layers, at the same time improving performances. The last iteration of the VB routing is also called \textit{ClassRouting} and it is used to route the highest-level information to the last layer of capsules before the feature space $\mathcal{F}$.

In our CapsNet, we employ four layers: a \textit{primary capsules} layer encapsulates the output features of $E$ into $16$-dimensional capsules, two \textit{convolutional capsules} layers refine the capsule features, and a final \textit{class capsules} layer encodes the output into a $J$-dimensional features in the latent space $\mathcal{F}$, where $J$ is the number of joints, also called $entities$.

Given each lower-level capsule $i$ and the corresponding higher-level capsule $j$, we define $M_{i}$ as the proposed lower level pose matrix and $W_{ij} \in \mathbb{R}^{4 \times 4}$ as a trainable viewpoint-equivariant transformation matrix such that:
    \begin{equation} 
    \label{eq:voting} 
    V_{j|i} = M_i W_{ij}
    \end{equation}
where $V_{j|i}$ is the vote coming from lower capsules $i$ for higher capsules $j$. The voting procedure takes place inside the VB routing and it allows each lower capsule $i$ to route its information to a higher capsule $j$ of its choice, thus allowing to build the hierarchical structure typical of CapsNets. 

To promote the viewpoint equivariance in Eq. \ref{eq:equivariance}, we introduce an inverse matrix $\hat{y}_{W}$ in the \textit{class capsules}, which aims at satisfying the Inverse Graphics constraint:

    \begin{equation} 
    \label{eq:inverse_graphics} 
    \hat{y}_{W} W_{ij} = I
    \end{equation}

 meaning that the learned inverse matrix $\hat{y}_{W}$ effectively acts as an approximated inverse of the rendering operation, as it is commonly found in computer graphics \cite{hinton2011transforming}.
 
At the output of the encoder, each \textit{entity} corresponding to each joint of the skeleton is defined by a flattened vector of $16$ elements, or, in other words, a $4\times4$ matrix, which is sufficient to grasp the complete pose (translation + rotation) of each joint. 

An overview of the capsule encoder is shown in Algorithm \ref{algo:encoder}. In the algorithm, $s_{3\mathcal{D}}, s_{2\mathcal{D}}, s_{\mathcal{DM}}, s_{\mathcal{W}}$ are weights used for the self-balancing of the loss, $w_c$ are the convolutional layer weights, $a$ are the activations of each Capsule layer, and $\{\cdot\}$ represents parameters used only when in the RGB domain. 

\begin{algorithm}
  \caption{Capsule encoder}
  \SetKwInOut{Input}{inputs}
  \SetKwInOut{Output}{outputs}
  \SetKwProg{CapsuleEncoder}{CapsuleEncoder}{}{}

  \CapsuleEncoder{$(x)$}{
    \Input{$x =  x_0 \dots x_{BS}$, ${BS} = $ batch size of RGB or depth images}
    \Output{
    $\mathcal{F} = J$ 16-dimensional $entities$; $\hat{y}_{W} =$ trainable Inverse Graphics matrix
    }
    $s_{3\mathcal{D}}, s_{2\mathcal{D}}, \{s_{\mathcal{DM}}\}, s_{\mathcal{W}} \gets 1$\;
    $w_c \gets xavier_{uniform}() \quad \forall c \in ConvLayers$\;
    \ForEach{$i \in ConvLayers$}{%
      $x \gets Conv2d_i(x)$\;
      $x \gets InstanceNorm2d_i(x)$\;
      $x \gets \text{GELU}(x)$\;
    }
    $a, x \gets PrimaryCapsules(x)$\;
    \ForEach{$j \in ConvCapsuleLayers$}{%
      $a, x \gets ConvCapsules_j(a, x)$\;
      $a, x \gets VBRouting_j(a, x)$\;
    }
    $a, x, \hat{y}_{W} \gets ClassCapsules(a, x)$\;
    $a, x \gets ClassRouting(a, x)$\;
    $\mathcal{F} \gets entities(x)$\;
    
    \KwRet{$\mathcal{F}, \hat{y}_{W}$}\;
  }
 \label{algo:encoder}
\end{algorithm}

\subsection{Multi-task decoders}
\label{subsec:decoders}

Starting from the 16-dimensional entities in the capsule feature space $\mathcal{F}$, we design a decoding module (light orange block in Fig. \ref{fig:network}) that allows us to simultaneously retrieve multiple predictions for different tasks from the same feature space $\mathcal{F}$.
Each decoder $D_{\tau}$ in the decoding module is configured as an independent fully connected block, with $0.5$ Dropout and GELU activations \cite{hendrycks2016gaussian}. We employ no weight sharing or layer sharing across the decoders to enforce the multi-task loss, as explained in section \ref{subsec:loss}. 

We define different tasks ($\tau$) with different objectives:

\begin{itemize}
    \item $3\mathcal{D}$: minimise the distance between ground truth and predicted 3D joints in 3D space $\hat{y}_{3D}$;
    \item $2\mathcal{D}$: as above, but without relying on 3D joints predictions, and rather predicting 2D joints $\hat{y}_{2D}$ as seen from the current viewpoint in camera frame coordinates;
    \item $\mathcal{DM}$: reconstruct the depth map $\hat{y}_{DM}$ of the input RGB image. It is used only in the RGB domain;
    \item $\mathcal{W}$ Inverse Graphics loss : learn the inverse graphics matrix $\hat{y}_{\mathcal{W}}$ to promote the \textit{de-rendering} of input pixels into isolated capsule entities, as explained in Sec. \ref{subsec:encoder}, Eq. \ref{eq:inverse_graphics}.
\end{itemize}

For each task $\tau = 3\mathcal{D}, 2\mathcal{D}, \mathcal{DM}$, a decoder $D_{\tau}$ takes as input the feature space $\mathcal{F}$ and it outputs the predictions $\hat{Y} = [\hat{y}_{3\mathcal{D}}, \hat{y}_{2\mathcal{D}}, \{\hat{y}_{\mathcal{DM}}\}]$ to the loss function. For $\mathcal{W}$, the $\hat{y}_{\mathcal{W}}$ matrix is forwarded to the loss function directly from the encoder.

An overview of the capsule decoders is shown in Algorithm \ref{algo:decoders}. 

\begin{algorithm}
  \caption{Capsule decoders}
  \SetKwInOut{Input}{inputs}
  \SetKwInOut{Output}{outputs}
  \SetKwProg{CapsuleDecoders}{CapsuleDecoders}{}{}

  \CapsuleDecoders{$(x)$}{
    \Input{$\mathcal{F} = J$ 16-dimensional $entities$}
    \Output{
    $\hat{Y} = [\hat{y}_{3\mathcal{D}}, \hat{y}_{2\mathcal{D}}, \{\hat{y}_{\mathcal{DM}}\}]$
    }
    $x \gets \mathcal{F}$;\\
    \ForEach{$i \in Y$}{%
      $x \gets Dropout_{0.5}(x)$\;
      $x \gets Linear(x)$\;
      $\hat{y}_{i} \gets \text{GELU}(x)$\;
    }
    
    \KwRet{$\hat{Y} = [\hat{y}_{3\mathcal{D}}, \hat{y}_{2\mathcal{D}}, \{\hat{y}_{\mathcal{DM}}\}]$}\;
  }
 \label{algo:decoders}
\end{algorithm}

\subsection{Self-balancing multi-task loss}
\label{subsec:loss}

Tasks are associated to the different input domains, as follows: 

\begin{table}[!htbp]
\centering
\resizebox{0.3\textwidth}{!}{%
\begin{tabular}{@{}lllll@{}}
\toprule
 & $3\mathcal{D}$ & $2\mathcal{D}$ & $\mathcal{DM}$ & $\mathcal{W}$ \\ \midrule
\rowcolor[HTML]{E2EFDA}
Depth & \checkmark   & \checkmark   &                           & \checkmark  \\
RGB   & \checkmark   & \checkmark   & \checkmark & \checkmark  \\ \bottomrule
\end{tabular}%
}
\label{tab:tasks}
\end{table}

Each task is assigned a loss $\mathcal{L}_{\tau}$, defined as:

\newcommand{\norm}[1]{\left\lVert#1\right\rVert}
\begin{itemize}
    \item $\mathcal{L}_{2\mathcal{D}}$, $\mathcal{L}_{3\mathcal{D}}$: Mean Square Error (MSE) loss for the $3\mathcal{D}$ and $2\mathcal{D}$ joints prediction tasks.
    \item $\mathcal{L}_{\mathcal{DM}}$: masked L1 loss for the depth estimation task $\mathcal{DM}$, in the RGB domain, where $mask$ is a function that applies the L1 loss only on pixels over a certain depth threshold, to promote the depth estimation over non-background areas.
    \item $\mathcal{L}_{\mathcal{W}}$: inverse graphics loss $\mathcal{W}$, which role is to enforce invertibility for the capsule weight matrices. The notation $\norm{.}_{F}$ defines the Frobenius norm of a matrix.
\end{itemize}

\begin{equation}
\label{eq:losses} 
\centering
    \mathcal{L}_{\tau} = 
    \begin{cases}
    \mathcal{L}_{2\mathcal{D},3\mathcal{D}} = \frac{1}{BS} \sum_{i=0}^{BS}(y_i - \hat{y_i})^2\\\\
    
    \mathcal{L}_{\mathcal{DM}} =\displaystyle\frac{\sum_{i=0}^{BS}\Big[ mask\displaystyle\left\lvert {y_i - \hat{y}_i} \right\rvert + \displaystyle\left\lvert {y_i - \hat{y}_i} \right\rvert\Big]}{\displaystyle 2 * BS}\\\\
    
    \mathcal{L}_{\mathcal{W}} = \norm{\hat{y}_{W} W_{ij}}_{F}
    \end{cases}
\end{equation}

Considering $\mathcal{T}$ as the set of the employed tasks ${\tau}$, the overall balanced loss for all the tasks is expressed as:

\begin{equation} 
\label{eq:loss} 
\centering
    \begin{split}
    \mathcal{L} &= \sum_{\tau \in \mathcal{T}} \left( s_{\tau} + e^{-s_{\tau}}\mathcal{L}_{\tau}\right) 
    \end{split}
\end{equation}


where $s_{\tau} = [s_{3\mathcal{D}}, s_{2\mathcal{D}}, s_{\mathcal{DM}}, s_{\mathcal{W}}]$ are the trainable weights associated with each loss in $\mathcal{T}$, initialised to 1 in algorithm \ref{algo:encoder}, and $\mathcal{L}_{\tau}$ is each loss of the enabled decoders, as defined in Eq. \ref{eq:losses}.

\section{Experiments}
\label{sec:experiments}

\subsection{Datasets}

\textbf{ITOP Dataset of depth images.} The ITOP dataset \cite{haque2016towards} contains depth images from top and front view. The training split and the test split consist of 40k and 10k images, respectively. The depth images display 15 videos of 20 actors in a constrained setting. The dataset is recorded using two Axus Xtion Pro cameras. The 3D skeleton model consists of 15 joints.

\textbf{PanopTOP31K dataset of depth and RGB images.} The PanopTOP dataset \cite{garau2021panoptop} consists of 31k top-view and 31k front view images coming from video sequences of 24 different actors, available both in the RGB and depth domain, for a total of 68k images. The ground truth 3D skeleton consists of 19 joints.

\textbf{Human3.6M dataset of RGB images.} For training and evaluation we follow the default protocol \#1 from Human3.6M \cite{ionescu2014human}, by reserving subjects $9$ and $11$ for evaluation, while only training on data from subjects $1, 5, 6, 7, 8$. Our architecture is fully end-to-end, requiring as input just one image and no additional information such as 2D joints ground truth, multiple sequential frames, or non-standard data-augmentation. Compared to the majority of methods present in literature, we don't rely on additional datasets for training, at the same time showing high generalization capabilities even after training on a subset of the possibly available data. 

\subsection{Evaluation metrics}

Following the works of \cite{haque2016towards,moon2018v2v,xiong2019a2j}, we choose the mean average precision (mAP) as the evaluation metric for the depth domain. It is defined as the percentage of all predicted joints which fall in an interval smaller than 0.10 meters. In the RGB domain, we use the Mean Per Joint Position Error (MPJPE) in millimeters as in many HPE works \cite{cao2017realtime,tome2017lifting,ramirez2020bayesian}.

\begin{figure*}[]
        \begin{subfigure}[b]{0.25\textwidth}
                \includegraphics[width=\linewidth]{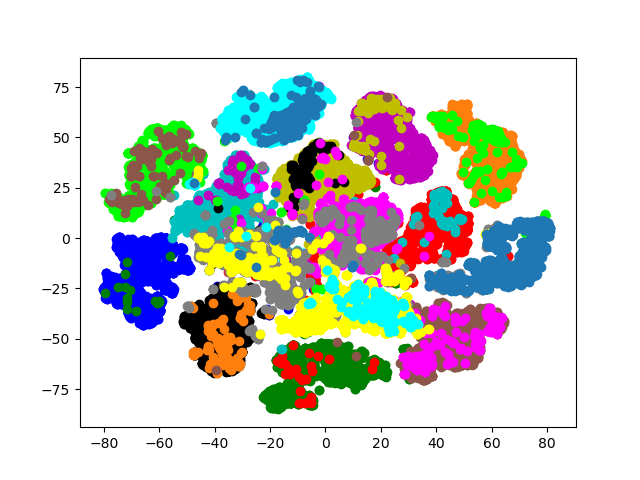}
                \caption{V2V \cite{moon2018v2v}}
                \label{fig:latentV2V}
        \end{subfigure}%
        \begin{subfigure}[b]{0.25\textwidth}
                \includegraphics[width=\linewidth]{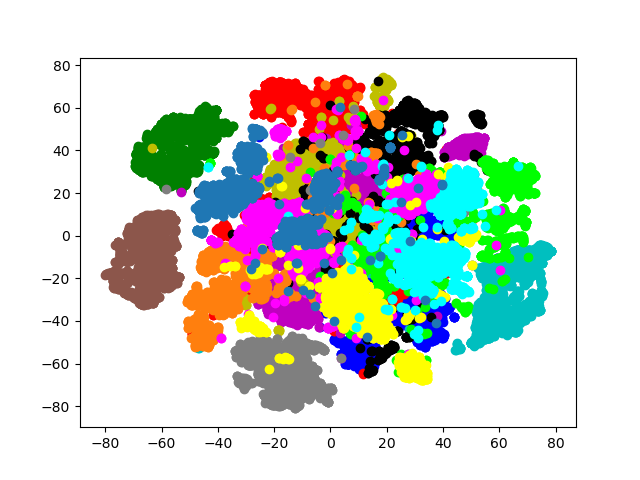}
                \caption{DECA-D1, $\mathcal{T} = [3\mathcal{D}]$}
                \label{fig:latent3D}
        \end{subfigure}%
        \begin{subfigure}[b]{0.25\textwidth}
                \includegraphics[width=\linewidth]{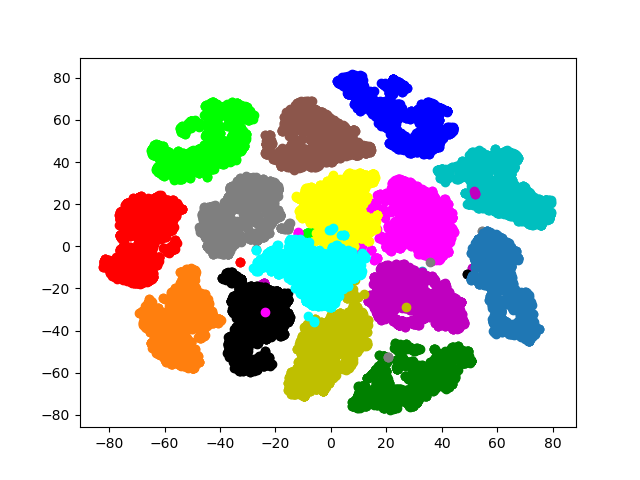}
                \caption{DECA-D2, $\mathcal{T} = [3\mathcal{D}, \mathcal{W}]$}
                \label{fig:latent3D+W}
        \end{subfigure}%
        \begin{subfigure}[b]{0.25\textwidth}
                \includegraphics[width=\linewidth]{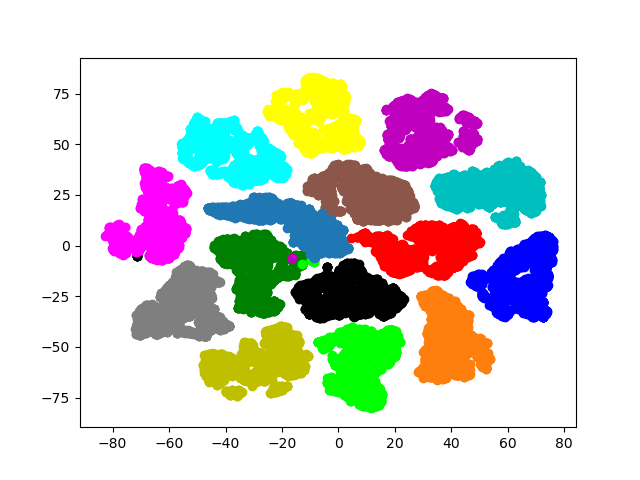}
                \caption{DECA-D3, $\mathcal{T} = [3\mathcal{D}, 2\mathcal{D}, \mathcal{W}]$}
                \label{fig:latent3D+2D+W}
        \end{subfigure}%
        \caption{2D representation on the 16-dimensional latent space obtained using t-SNE \cite{van2008visualizing}. Each dot corresponds to an entity $E_{jt}$ representing a joint $jt$ of the skeleton from the test set of ITOP \cite{haque2016towards}. V2V network \cite{moon2018v2v} relies on CNNs, thus is not able to cluster together samples corresponding to the same entity (a). When trained to satisfy only the 3D prediction constraint our DECA-D1 network performs slightly better than V2V (b). The 15 clusters, corresponding to the 15 joints of the skeleton model, are clearly distinguishable in DECA-D2 (c) and DECA-D3 (d), with (d) displaying better cluster separation and fewer outliers.}\label{fig:latent2D}
\end{figure*}

\begin{figure}[ht]
    \centering
    \includegraphics[width=180pt, height=150pt]{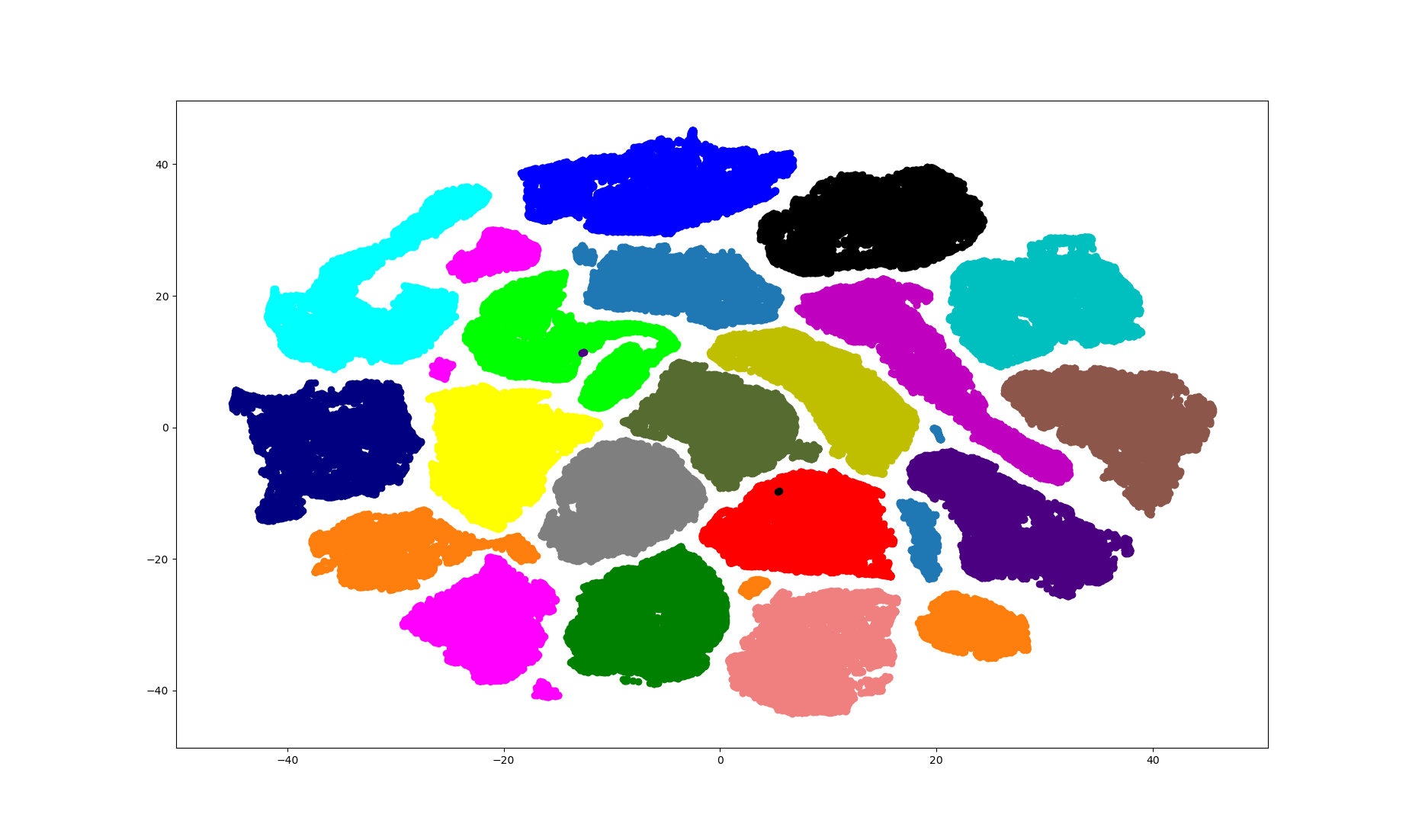}
    \caption{Organization of the latent space of DECA-H4 after t-SNE for the Human3.6M dataset: the color of each sample point corresponds to each joint class.}
    \label{fig:latent_human}
\end{figure}

\subsection{Implementation details}

Our network is trained in an end-to-end fashion using Pytorch Lightning. Input images are normalized in the interval $[0,1]$ with a resolution of 256x256 pixels for depth images and 256x256 pixels for RGB ones. We do not perform any augmentations on the input datasets. The batch size is set to 128 for ITOP and PanopTOP31K and to 32 for Human3.6M. We initialize the weights with the Xavier initialization \cite{glorot2010understanding}. The learning rate is set to $1e^{-5}$, the weight decay is set to 0, and Adam is the optimizer of choice. We train our network for approximately 20 epochs for each dataset.

\subsection{Feature space entities and ablation study}
\label{sec:ablation}

We report experiments on the top-view of the ITOP dataset \cite{haque2016towards} to validate the 3D representation provided by our network and to show how the multi-tasks decoder influences the overall performances. 

To do so, we deploy 5 configurations, 3 on depth data and 2 on RGB data, with different sets of tasks $\mathcal{T}$ of our method:
\begin{itemize}
    \item DECA-D1, with $\mathcal{T} = [3\mathcal{D}]$
    \item DECA-D2, with $\mathcal{T} = [3\mathcal{D}, \mathcal{W}]$
    \item DECA-D3, with $\mathcal{T} = [3\mathcal{D}, 2\mathcal{D}, \mathcal{W}]$
    \item DECA-R4, with $\mathcal{T} = [3\mathcal{D}, 2\mathcal{D}, \mathcal{DM}, \mathcal{W}]$
    \item DECA-H4, with $\mathcal{T} = [3\mathcal{D}, 2\mathcal{D}, \mathcal{DM_{J}}, \mathcal{W}]$
\end{itemize}


where the letter $D$ or $R$ indicates the depth or RGB domains, and the number defines how many tasks are assigned to the network.
Since we are evaluating the performances on the 3D HPE, the $\tau = [3\mathcal{D}]$ is used for all the different configurations. In the case of $DECA_H4$ (trained on Human3.6M), $DM_{J}$ refers to the task of estimating the  joint heatmaps instead of the image depth maps $DM$ as in $DECA_R4$.

\textbf{Loss effectiveness analysis}. 
The results are reported in the last 3 columns of Table \ref{tab:results_itop_top}. As shown in the Table, increasing the number of tasks in $\mathcal{T}$ generally leads to an increase in the network's performances. DECA-D1 already achieves similar results to the state-of-the-art, thanks to the CapsNets' capability to interpret the geometrical nature of the input data. When the inverse graphics loss $\mathcal{W}$ is employed (DECA-D2 and DECA-D3), the enforced invertibility of the weights matrix leads to an immediate gain in performances. 
In DECA-D3, the introduction of the $2\mathcal{D}$ loss leads to an additional improvement in terms of accuracy. Hence, we argue that the network performances improve when more tasks are given because we achieve a better representation of the entities in the latent space.


\textbf{Latent space analysis}. 
To analyze the latent space, we use the features of the test set extracted after the capsule modules. Each feature $f \in \mathcal{F}$ is linearised to obtain a vector of length $L_{feat}$.  At this stage, each entity $E_{jt}$ corresponding to each joint $jt$ is defined by dividing each feature vector by the number of joints, resulting in vectors of length $\frac{L_{feat}}{\# of joints}$. For visualisation purposes, we use \textit{t-SNE} \cite{van2008visualizing} to project the entities on a 2-dimensional space. The results are displayed in Fig. \ref{fig:latent2D}. We compare our latent space against the publicly available version of the V2V \cite{moon2018v2v} encoder/decoder structure. We show how our DECA network can better cluster and separate each entity $E_{jt}$ with respect to V2V. Our solution provides a better organization of the latent space, with bigger inter-class margins and fewer outliers. The latent space organization improves drastically when we employ the $\tau=\mathcal{W}$ task (DECA-D2), thus enforcing the inverse graphics constraint. In DECA-D3 we add the $\tau=2\mathcal{D}$ task. The resulting organization of the latent space improves, thus further establishing a correlation between the growing number of tasks and the improvement in performances.

\begin{table*}[]
\centering
\resizebox{.8\textwidth}{!}{%
\begin{tabular}{@{}lcccccccc@{}}
\toprule
           & \multicolumn{8}{c}{\textbf{ITOP front-view}}                                                   
           \\ \midrule
Body part  & RF\cite{shotton2011real}    & RTW\cite{yub2015random}            & IEF\cite{carreira2016human}   & VI \cite{haque2016towards}            & REN9x6x6\cite{guo2017towards}       & V2V\cite{moon2018v2v}            & A2J\cite{xiong2019a2j}   & DECA-D3        \\ \midrule
\rowcolor[HTML]{E2EFDA}
Head       & 63.80 & 97.80          & 96.20 & 98.10          & \textbf{98.70} & 98.29          & 98.54 & 93.87          \\
Neck       & 86.40 & 95.80          & 85.20 & 97.50          & \textbf{99.40} & 99.07          & 99.20 & 97.90          \\
\rowcolor[HTML]{E2EFDA}
Shoulders  & 83.30 & 94.10          & 77.20 & 96.50          & 96.10          & \textbf{97.18} & 96.23 & 95.22          \\
Elbows     & 73.20 & 77.90          & 45.40 & 73.30          & 74.70          & 80.42          & 78.92 & \textbf{84.53} \\
\rowcolor[HTML]{E2EFDA}
Hands      & 51.30 & \textbf{70.50} & 30.90 & 68.70          & 55.20          & 67.26          & 68.35 & 56.49          \\
Torso      & 65.00 & 93.80          & 84.70 & 85.60          & 98.70          & 98.73          & 98.52 & \textbf{99.04} \\
\rowcolor[HTML]{E2EFDA}
Hip        & 50.80 & 80.30          & 83.50 & 72.00          & 91.80          & 93.23          & 90.85 & \textbf{97.42} \\
Knees      & 65.70 & 68.80          & 81.80 & 69.00          & 89.00          & 91.80          & 90.75 & \textbf{94.56} \\
\rowcolor[HTML]{E2EFDA}
Feet       & 61.30 & 68.40          & 80.90 & 60.80          & 81.10          & 87.60          & 86.91 & \textbf{92.04} \\
Upper Body & -     & -              & -     & \textbf{84.00} & -              & -              & -     & 83.03          \\
\rowcolor[HTML]{E2EFDA}
Lower Body & -     & -              & -     & 67.30          & -              & -              & -     & \textbf{95.30} \\
\rowcolor[HTML]{FFFFC7}
Mean       & 65.80 & 80.50          & 71.00 & 77.40          & 84.90          & 88.74          & 88.00 & \textbf{88.75} \\ \bottomrule
\end{tabular}%
}
\caption{Comparison with the state-of the art for ITOP front-view (metric: 0.1m mAP).}
\label{tab:results_itop_front}
\end{table*}

\begin{table*}[]
\centering
\resizebox{.8\textwidth}{!}{%
\begin{tabular}{@{}lcccccccccc@{}}
\toprule
          & \multicolumn{10}{c}{\textbf{ITOP top-view}}                              
          \\ \midrule
Body part  & RF\cite{shotton2011real}    & RTW\cite{yub2015random}            & IEF\cite{carreira2016human}   & VI \cite{haque2016towards}            & REN9x6x6\cite{guo2017towards}       & V2V\cite{moon2018v2v}            & A2J\cite{xiong2019a2j}    & DECA-D1 & DECA-D2        & DECA-D3        \\ \midrule
\rowcolor[HTML]{E2EFDA}
Head       & 95.40 & \textbf{98.40} & 83.80 & 98.10          & 98.20    & \textbf{98.40} & 98.38 & 94.41   & 95.31          & 95.37          \\
Neck       & 98.50 & 82.20          & 50.00 & 97.60          & 98.90    & 98.91          & 98.91 & 98.86   & \textbf{99.16} & 98.68          \\
\rowcolor[HTML]{E2EFDA}
Shoulders  & 89.00 & 91.80          & 67.30 & 96.10          & 96.60    & 96.87          & 96.26 & 96.12   & \textbf{97.51} & 96.57          \\
Elbows     & 57.40 & 80.10          & 40.20 & \textbf{86.20} & 74.40    & 79.16          & 75.88 & 76.86   & 81.67          & 84.07          \\
\rowcolor[HTML]{E2EFDA}
Hands      & 49.10 & 76.90          & 39.00 & \textbf{85.50} & 50.70    & 62.44          & 59.35 & 44.41   & 45.97          & 54.33          \\
Torso      & 80.50 & 68.20          & 30.50 & 72.90          & 98.10    & 97.78          & 97.82 & 99.46   & \textbf{99.70} & 99.46          \\
\rowcolor[HTML]{E2EFDA}
Hip        & 20.00 & 55.70          & 38.90 & 61.20          & 85.50    & 86.91          & 86.88 & 97.84   & \textbf{97.87} & 97.42          \\
Knees      & 2.60  & 53.90          & 54.00 & 51.60          & 70.00    & 83.28          & 79.66 & 88.01   & 88.19          & \textbf{90.84} \\
\rowcolor[HTML]{E2EFDA}
Feet       & 0.00  & 28.70          & 62.40 & 51.50          & 41.60    & 69.62          & 58.34 & 79.30   & \textbf{83.53} & 81.88          \\
Upper Body & -     & -              & -     & \textbf{91.40} & -        & -              & -     & 78.51   & 80.60          & 83.00          \\
\rowcolor[HTML]{E2EFDA}
Lower Body & -     & -              & -     & 54.70          & -        & -              & -     & 89.96   & 91.27          & \textbf{91.39} \\
\rowcolor[HTML]{FFFFC7}
Mean       & 47.40 & 68.20          & 51.20 & 75.50          & 75.50    & 83.44          & 80.5  & 83.85   & 85.58          & \textbf{86.92} \\ \bottomrule
\end{tabular}%
}
\caption{Comparison with the state-of the art for ITOP top-view (metric: 0.1m mAP).}
\label{tab:results_itop_top}
\end{table*}

\subsection{Comparison with state-of-the-art methods}

\textbf{Depth data: ITOP dataset.} We compare our DECA against common state-of-the-art method for human pose estimation on depth images \cite{shotton2011real, yub2015random, carreira2016human, haque2016towards, guo2017towards, moon2018v2v, xiong2019a2j}. The results are reported in Tab. \ref{tab:results_itop_front} and Tab. \ref{tab:results_itop_top}. 
Our DECA outperforms existing methods on the front-view task, improving the accuracy by a wide margin on the more challenging top viewpoint. In general, we consistently perform better than other methods on most of the joints and the average. The gain of our method is particularly large when dealing with the lower body, which is often occluded in the top-view.

\textbf{Depth data: Viewpoint-equivariant ITOP.} We test DECA on the viewpoint transfer task, meaning training on one viewpoint, either top-view or front-view, and testing on the other one, unseen at training time. The comparison against available state-of-the-art methods \cite{shotton2011real, yub2015random, carreira2016human, haque2016towards} are reported in Tab. \ref{tab:viewpoint_transfer_ITOP}. We consistently outperform other methods by a wide margin, thus making a step forward toward viewpoint equivariance. 
While other methods provide only the best subset of viewpoint transfer results (Tab. \ref{tab:viewpoint_transfer_ITOP}), omitting entirely the train on top and test on front scenario, we provide results for all the joints and all the viewpoint transfer combinations in Tab. \ref{tab:viewpoint_transfer_ITOP_both}. Our DECA achieves better results than the top-most of the other methods on many different joints (e.g. shoulders, lower body). In Tab \ref{tab:viewpoint_transfer_ITOP_both}, training DECA on top-view or front-view achieves comparable lower body accuracy. This means that when the network is trained on top view, where the lower body is mostly occluded, it can retrieve the occluded joints from previously unseen front views, and vice versa. This shows how our network has learned the viewpoint as a parameter, and it is thus able to generalize in a similar fashion in all the viewpoint transfer combinations. 

\begin{table}[]
\centering
\resizebox{.45\textwidth}{!}{%
\begin{tabular}{@{}lccccc@{}}
\toprule
           & \multicolumn{5}{c}{\textbf{\begin{tabular}[c]{@{}c@{}}ITOP\\Train on front, test on top\end{tabular}}} \\ \midrule
Body part  & RF \cite{shotton2011real}                    & RTW \cite{yub2015random}                  & IEF \cite{carreira2016human}                   & VI \cite{haque2016towards}                    & DECA-D3                       \\ \midrule
\rowcolor[HTML]{E2EFDA}
Head       & 48.10                 & 1.50                 & 47.90                 & 55.60                 & 46.27                          \\
Neck       & 5.90                  & 8.10                 & 39.00                 & 40.90                 & \textbf{73.14}                 \\
\rowcolor[HTML]{E2EFDA}
Torso      & 4.70                  & 3.90                 & 41.90                 & 35.00                 & \textbf{85.94}                 \\
Upper Body & 19.70                 & 2.20                 & 23.90                 & 29.40                 & \textbf{45.00}                 \\
\rowcolor[HTML]{FFFFC7} 
Full Body  & 10.80                 & 2.00                 & 17.40                 & 20.40                 & \textbf{51.85}                 \\ \bottomrule
\end{tabular}%
}
\vspace*{7pt}
\caption{Comparison with the state-of the art for the ITOP viewpoint transfer task (metric: 0.1m mAP). Training on front-view, validating on front-view, testing on top-view (top-view data is unseen in validation).}
\label{tab:viewpoint_transfer_ITOP}
\end{table}

\begin{table}[]
\centering
\resizebox{.3\textwidth}{!}{%
\begin{tabular}{@{}lcc@{}}
\toprule
           & \multicolumn{2}{c}{\textbf{DECA-D3}}                                                                                                                                 \\ \midrule
Body part  & \begin{tabular}[c]{@{}c@{}}Train on front,\\ test on top\end{tabular} & \begin{tabular}[c]{@{}c@{}}Train on top,\\ test on front\end{tabular} \\ \midrule
\rowcolor[HTML]{E2EFDA}
Head       & 46.27                                                                           & 18.51                                                                              \\
Neck       & 73.14                                                                           & 44.77                                                                              \\
\rowcolor[HTML]{E2EFDA}
Shoulders  & 69.02                                                                           & 25.18                                                                              \\
Elbows     & 43.87                                                                           & 16.23                                                                              \\
\rowcolor[HTML]{E2EFDA}
Hands      & 9.41                                                                            & 2.19                                                                               \\
Torso      & 85.94                                                                           & 68.63                                                                              \\
\rowcolor[HTML]{E2EFDA}
Hip        & 72.15                                                                           & 64.75                                                                              \\
Knees      & 49.31                                                                           & 68.15                                                                              \\
\rowcolor[HTML]{E2EFDA}
Feet       & 42.46                                                                           & 46.12                                                                              \\
Upper Body & 45.00                                                                           & 18.81                                                                              \\
\rowcolor[HTML]{E2EFDA}
Lower Body & 59.11                                                                           & 60.95                                                                              \\
\rowcolor[HTML]{FFFFC7} 
Mean       & 51.85                                                                           & 38.48                                                                              \\ \bottomrule
\end{tabular}%
}
\vspace*{7pt}
\caption{DECA-D3 complete results for the ITOP viewpoint transfer tasks (metric: 0.1m mAP). Test data is unseen during validation for both the cases.}
\label{tab:viewpoint_transfer_ITOP_both}
\end{table}

\textbf{RGB data: Viewpoint-equivariant PanopTOP31K.} 
To the best of our knowledge, we are the first to tackle the problem of viewpoint transfer between top-view and front-view in the RGB domain. We report results with training and testing on both seen and unseen viewpoints in Tab. \ref{tab:results_rgb}. The chosen metric is the mean per-joint projection error (MPJPE). We report results with and without the Procrustes alignment \cite{goodall1991procrustes} of the predicted poses. It is interesting to notice how DECA can reduce the gap between the same viewpoint results and the results of the viewpoint transfer tasks. In the case of viewpoint transfer, we train on viewpoint A, validate on the same viewpoint A and test on viewpoint B.

\begin{table*}[]
\centering
\resizebox{.89\textwidth}{!}{%
\begin{tabular}{@{}lcccccccc@{}}
\toprule
            & \multicolumn{8}{c}{\textbf{DECA-R4}}                                                                                                                                                                                                                                                                                                                                                                              \\ \midrule
            & \multicolumn{2}{c}{\textbf{\begin{tabular}[c]{@{}c@{}}Train on front,\\ test on front\end{tabular}}} & \multicolumn{2}{c}{\textbf{\begin{tabular}[c]{@{}c@{}}Train on top,\\ test on top\end{tabular}}} & \multicolumn{2}{c}{\textbf{\begin{tabular}[c]{@{}c@{}}Train on front,\\ test on top\end{tabular}}} & \multicolumn{2}{c}{\textbf{\begin{tabular}[c]{@{}c@{}}Train on top,\\ test on front\end{tabular}}} \\ \midrule
Body part   & No Procrustes                                      & Procrustes                                      & No Procrustes                                    & Procrustes                                    & No Procrustes                                     & Procrustes                                     & No Procrustes                                     & Procrustes                                     \\ \midrule
\rowcolor[HTML]{E2EFDA} 
Neck        & 4.02                                               & 2.37                                            & 4.55                                             & 2.51                                          & 16.02                                             & 4.16                                           & 8.21                                              & 5.06                                           \\
Nose        & 5.66                                               & 3.75                                            & 6.98                                             & 3.89                                          & 16.83                                             & 7.67                                           & 10.72                                             & 6.76                                           \\
\rowcolor[HTML]{E2EFDA} 
Body Center & 0.56                                               & 4.63                                            & 1.23                                             & 3.63                                          & 1.01                                              & 31.20                                          & 0.83                                              & 11.59                                          \\
Shoulders   & 4.56                                               & 2.76                                            & 5.14                                             & 3.07                                          & 17.43                                             & 5.33                                           & 8.51                                              & 5.35                                           \\
\rowcolor[HTML]{E2EFDA} 
Elbows      & 9.82                                               & 7.14                                            & 9.64                                             & 7.51                                          & 29.70                                             & 18.52                                          & 23.20                                             & 15.47                                          \\
Hands       & 13.88                                              & 10.82                                           & 14.02                                            & 12.34                                         & 47.01                                             & 38.29                                          & 36.78                                             & 28.25                                          \\
\rowcolor[HTML]{E2EFDA} 
Hips        & 18.75                                              & 4.87                                            & 2.71                                             & 3.89                                          & 5.10                                              & 30.07                                          & 3.64                                              & 10.88                                          \\
Knees       & 9.54                                               & 5.14                                            & 7.59                                             & 4.84                                          & 52.98                                             & 28.65                                          & 20.11                                             & 9.28                                           \\
\rowcolor[HTML]{E2EFDA} 
Feet        & 11.53                                              & 5.08                                            & 9.83                                             & 5.10                                          & 69.18                                             & 28.75                                          & 26.36                                             & 11.07                                          \\
Eyes        & 6.19                                               & 4.00                                            & 7.44                                             & 3.79                                          & 19.33                                             & 11.00                                          & 11.40                                             & 7.45                                           \\
\rowcolor[HTML]{E2EFDA} 
Ears        & 5.50                                               & 3.73                                            & 7.15                                             & 3.74                                          & 23.56                                             & 13.00                                          & 11.22                                             & 7.16                                           \\
Upper Body  & 6.93                                               & 5.21                                            & 7.66                                             & 5.46                                          & 23.69                                             & 16.56                                          & 15.54                                             & 11.60                                          \\
\rowcolor[HTML]{E2EFDA} 
Lower Body  & 7.65                                               & 5.03                                            & 6.71                                             & 4.61                                          & 42.42                                             & 29.16                                          & 16.71                                             & 10.41                                          \\
\rowcolor[HTML]{FFFFC7} 
Mean        & 7.16                                               & 5.15                                            & 7.36                                             & 5.19                                          & 29.60                                             & 20.54                                          & 15.91                                             & 11.22                                          \\ \bottomrule
\end{tabular}%
}
\caption{DECA-R4 results on the PanopTOP31K RGB dataset, with and without the Procrustes transformation  \cite{goodall1991procrustes} (metric: MPJPE). Tasks: (i) 3D pose estimation from the front and top viewpoints (ii) viewpoint transfer for both front and top views. Test data is unseen during validation for both the viewpoint transfer tasks.}
\label{tab:results_rgb}
\end{table*}

\begin{figure}[ht]
    \centering
    \includegraphics[width=.6\textwidth]{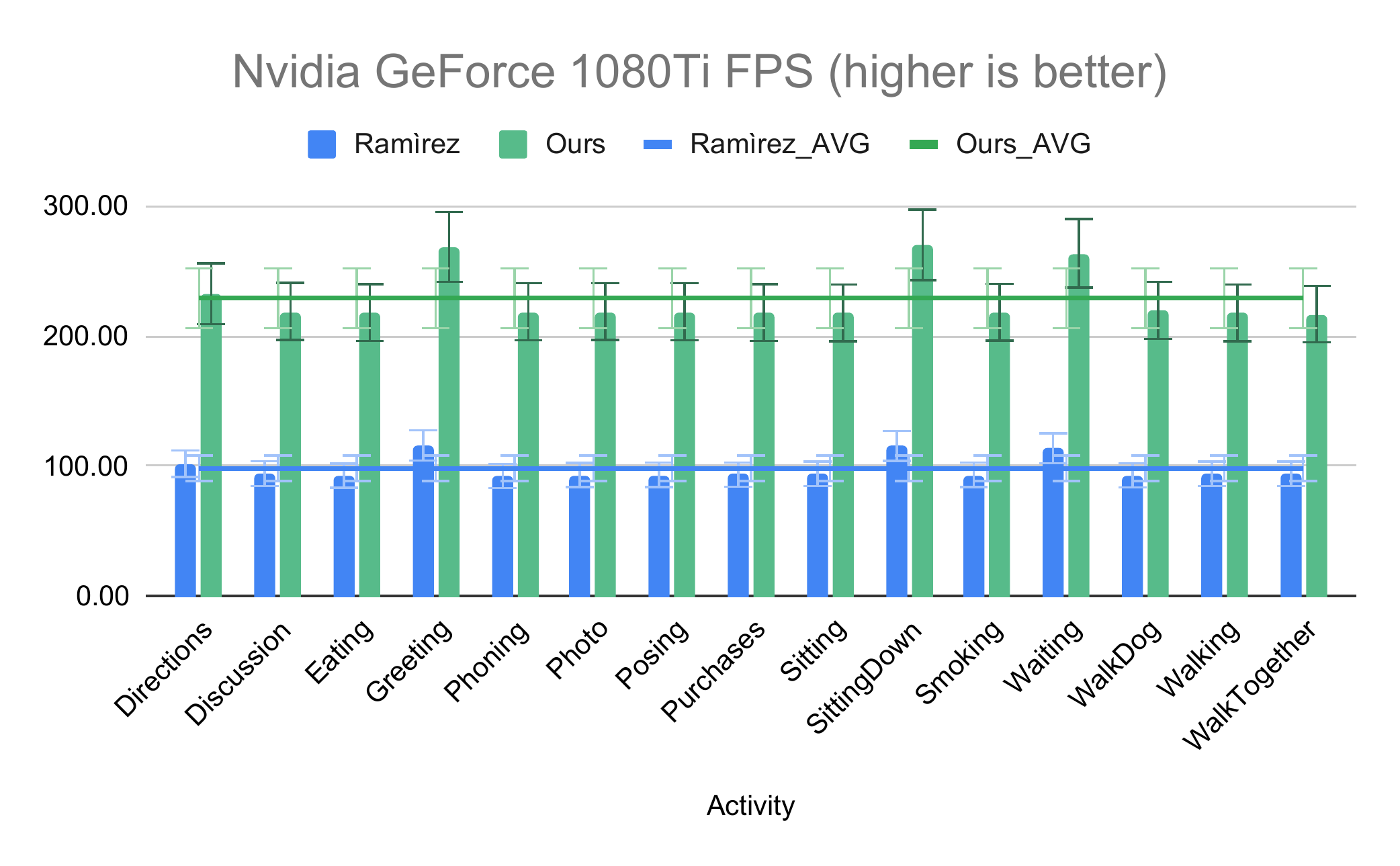}
    \caption{Activity-wise and average inference speed comparison on the same hardware (Nvidia GeForce 1080Ti).}
    \label{fig:1080Ti}
\end{figure}

\begin{figure}[ht]
    \centering
    \includegraphics[width=.6\textwidth]{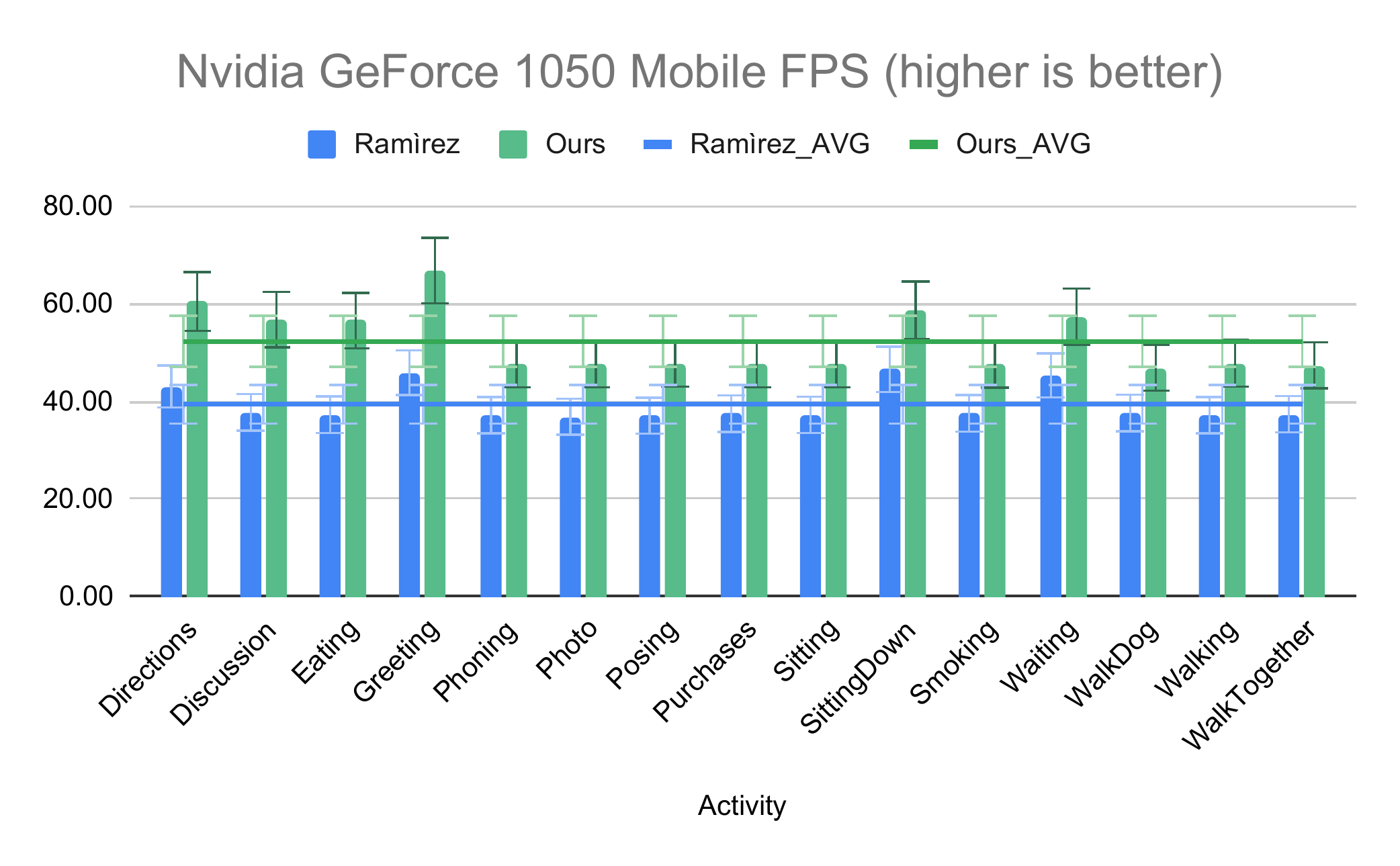}
    \caption{Activity-wise and average inference speed comparison on the same hardware (Nvidia GeForce 1050 Mobile).}
    \label{fig:1050}
\end{figure}

\textbf{RGB data: Viewpoint-equivariant Human3.6M.} 
Most works in literature try to achieve state-of-the-art results on Human3.6M with respect to joint accuracy, discarding the importance of viewpoint equivariance. As a result, they are usually not able to generalize with respect to unseen viewpoints (Fig. \ref{fig:teaser}). 
Our architecture is fully end-to-end, requiring as input just one image and no additional information such as 2D joints ground truth, multiple sequential frames, or non-standard data-augmentation. Compared to the majority of methods present in literature, we don't rely on additional datasets for training, at the same time showing high generalization capabilities even after training on a subset of the possibly available data. The metrics we use for comparison are the Mean Per Joint Position Error (MPJPE) in millimiters for each of the 15 activities in the Human3.6M dataset and the average by activity MPJPE, for each camera in the dataset. 
As for the implementation, the network we present is written using Pytorch Lightning, focusing on high modularity, allowing for real-time joint 3D and 2D predictions, achieving over $229$ FPS ($0.00436 s/frame$) on an Nvidia GeForce 1080Ti (desktop) and over $52$ FPS ($0.01913 s/frame$) on and Nvidia GeForce 1050 Mobile (laptop), almost twice as fast as what is reported in \cite{ramirez2020bayesian}. All the results were conducted on the same exact hardware and in the same conditions. In Fig. \ref{fig:1080Ti} we show the activity-wise and mean frames per second of our architecture compared to the other capsule-based networks \cite{ramirez2020bayesian} on a high-end, desktop-grade GPU. In this scenario, our architecture allows for a $2.33\times$ speed-up. Even in more resource-constrained scenarios (laptop-grade GPU, Fig. \ref{fig:1050}) we manage to gain an additional $\sim15 FPS$ on average. According to our experiments, the biggest improvements in terms of speed mostly come down to a combination of simplified network structure, the usage of the improved capsule paradigm and faster routing.
In the following sections we show some quantitative and qualitative results as well, both from the Human3.6 dataset and in-the-wild.

\subsection{Quantitative results}


In Table \ref{tab:results} we show our results compared to the state-of-the-art methods, both the ones using Procrustes transformation (right) and the ones reporting results without Procrustes (left). We achieve the lowest average MPJPE on both the categories and on most of the activities, without using additional information or non-standard data augmentation. Works using additional data, such as 2D-to-3D lifting, ground truth 2D joints, multiple datasets or temporal information are marked in Table \ref{tab:results} with a \textcolor{red}{*} symbol. We achieve similar or better results even with those methods, without relying on additional information, dataset or data augmentation, as shown in Table \ref{tab:recent}. Even considering other similar works that employ additional information, we obtain the lowest average MPJPE scores (yellow row). Compared to the only other work in literature using CapsNet \cite{ramirez2020bayesian}, our model achieves better MPJPE in almost every activity.\\
For the sake of completeness, we selected the top recent works in literature (2019-2020) with the lowest average MPJPE on the Human3.6M dataset, working on monocular data (Table \ref{tab:recent}). However, as Table \ref{tab:recent} shows, most of the works are aided by 2D ground truth information, meaning that they cannot be properly considered end-to-end. Additionally, many of them even exploit temporal frame sequences to refine joint predictions, thus non working with single images. Others use additional datasets, hand-crafted data augmentation of biometric models during training. We stress the fact that a big advantage of employing capsule networks is the increased generalization capabilities, which highly reduce the need for additional training data, and at the same time boosting network efficiency. Nonetheless, even considering the more recent results that use additional information or datasets, our results remain comparable.

\begin{table*}[]
\centering
\resizebox{\textwidth}{!}{%
\begin{tabular}{@{}lcccccccc|cccc@{}}
\toprule
{\color[HTML]{FFFFFF} \textbf{}}         & \multicolumn{8}{c|}{ \textbf{No Procrustes}}                                                                                                                                                                                                                                                                         & \multicolumn{4}{c}{\textbf{Procrustes}}                                                                                          \\

{ \textbf{Activity}} & { \makecell{\textbf{Zhou \textcolor{red}{*}} \\ \cite{zhou2016sparseness}}} & { \makecell{\textbf{Tekin \textcolor{red}{*}}\\ \cite{tekin2017learning}}}     & { \makecell{\textbf{Tome, I \textcolor{red}{*}}\\ \cite{tome2017lifting}}} & { \makecell{\textbf{Ramìrez, I} \\ \cite{ramirez2020bayesian}}} & { \makecell{\textbf{Tome, II \textcolor{red}{*}}\\ \cite{tome2017lifting}}}  & { \makecell{\textbf{Ramìrez, II} \\ \cite{ramirez2020bayesian}}} & { \makecell{\textbf{Ramìrez, III} \\ \cite{ramirez2020bayesian}}} & { \makecell{\textbf{DECA-H4, I}}} & { \makecell{\textbf{Sanzari \textcolor{red}{*}}\\ \cite{sanzari2016bayesian}}}   & { \makecell{\textbf{Bogo \textcolor{red}{*}}\\ \cite{bogo2016keep}}}   & { \makecell{\textbf{Ramìrez, IV} \\ \cite{ramirez2020bayesian}}} & { \makecell{\textbf{DECA-H4, II}}} \\ \toprule
\rowcolor[HTML]{E2EFDA} 
{ Directions}        & { 87.36}           & { 85.03}                & { 68.55}              & { 79.42}               & { {\ul \textit{64.98}}} & { 73.15}                & { 73.33}                 & { \textbf{70.16}}   & { {\ul \textit{48.82}}} & { 62}                & { 57.55}                & { \textbf{55.02}}    \\
{ Discussion}        & { 109.31}          & { 108.79}               & { 78.27}              & { 83.73}               & { {\ul \textit{73.47}}} & { 84.95}                & { 83.45}                 & { \textbf{76.67}}   & { {\ul \textit{56.31}}} & { 60.2}              & { 61.32}                & { \textbf{58.06}}    \\
\rowcolor[HTML]{E2EFDA} 
{ Eating}            & { 87.05}           & { 84.38}                & { 77.22}              & { 84.01}               & { {\ul \textit{76.82}}} & { 85.87}                & { 85.33}                 & { \textbf{78.41}}   & { 95.98}                & { 67.8}              & { 66.48}                & { \textbf{60.91}}    \\
{ Greeting}          & { 103.16}          & { 98.94}                & { 89.05}              & { 83.15}               & { 86.43}                & { 80.12}                & { 79.08}                 & { \textbf{76.87}}   & { 84.78}                & { 76.5}              & { 64.49}                & { \textbf{61.69}}    \\
\rowcolor[HTML]{E2EFDA} 
{ Phoning}           & { 116.18}          & { 119.39}               & { 91.63}              & { \textbf{86.42}}      & { {\ul \textit{86.28}}} & { 91.44}                & { 89.99}                 & { 87.99}            & { 96.47}                & { 92.1}              & { 68}                   & { \textbf{66.49}}    \\
{ Photo}             & { 143.32}          & { 95.65}                & { 110.05}             & { 112.38}              & { 110.67}               & { \textbf{109.42}}      & { 109.95}                & { 109.49}           & { 105.58}               & { {\ul \textit{77}}} & { 83.16}                & { \textbf{80.02}}    \\
\rowcolor[HTML]{E2EFDA} 
{ Posing}            & { 106.88}          & { 98.49}                & { 74.92}              & { 81.34}               & { {\ul \textit{68.93}}} & { 76.40}                & { 76.08}                 & { \textbf{72.23}}   & { 66.3}                 & { 73}                & { 56.05}                & { \textbf{54.94}}    \\
{ Purchases}         & { 99.78}           & { 93.77}                & { 83.71}              & { 77.65}               & { 74.79}                & { 76.72}                & { 73.61}                 & { \textbf{73.12}}   & { 107.41}               & { 75.3}              & { 54.85}                & { \textbf{52.89}}    \\
\rowcolor[HTML]{E2EFDA} 
{ Sitting}           & { 124.52}          & { {\ul \textit{73.76}}} & { 115.94}             & { 105.10}              & { 110.19}               & { 105.54}               & { \textbf{104.12}}       & { 108.84}           & { 116.89}               & { 100.3}             & { \textbf{77.65}}       & { 80.11}             \\
{ SittingDown}       & { 199.23}          & { 170.40}               & { 185.72}             & { 135.55}              & { 173.91}               & { \textbf{130.15}}      & { 136.27}                & { 149.53}           & { 129.63}               & { 137.3}             & { \textbf{97.32}}       & { 99.84}             \\
\rowcolor[HTML]{E2EFDA} 
{ Smoking}           & { 107.42}          & { 85.08}                & { 88.25}              & { 88.25}               & { {\ul \textit{84.95}}} & { 88.07}                & { 87.59}                 & { \textbf{87.29}}   & { 97.84}                & { 83.4}              & { \textbf{67.31}}       & { 67.86}             \\
{ Waiting}           & { 118.09}          & { 116.91}               & { 88.73}              & { 79.24}               & { 85.78}                & { 80.25}                & { 79.19}                 & { \textbf{75.14}}   & { 65.94}                & { 77.3}              & { 59.63}                & { \textbf{57.71}}    \\
\rowcolor[HTML]{E2EFDA} 
{ WalkDog}           & { 114.23}          & { 113.72}               & { 92.37}              & { 87.45}               & { {\ul \textit{86.26}}} & { 88.75}                & { \textbf{87.13}}        & { 87.70}            & { 130.46}               & { 79.7}              & { \textbf{64.76}}       & { 65.28}             \\
{ Walking}           & { 79.39}           & { 62.08}                & { 76.48}              & { 67.56}               & { 71.36}                & { \textbf{66.10}}       & { 66.31}                 & { \textbf{65.38}}   & { 92.58}                & { 86.8}              & { \textbf{49.96}}       & { 51.19}             \\
\rowcolor[HTML]{E2EFDA} 
{ WalkTogether}      & { 97.70}           & { 94.83}                & { 77.95}              & { 80.45}               & { {\ul \textit{73.14}}} & { 76.84}                & { 76.88}                 & { \textbf{75.76}}   & { 102.21}               & { 81.7}              & { \textbf{60.47}}       & { 61.04}             \\ \bottomrule
\rowcolor[HTML]{FFFFC7} 
{ Avg, by activity}  & { 112.91}          & { 100.08}               & { 93.26}              & { 88.78}               & { 88.53}                & { 87.58}                & { 87.22}                 & { \textbf{86.17}}   & { 93.15}                & { 82.03}             & { 65.93}                & { \textbf{64.98}}    \\
\rowcolor[HTML]{FFDEAD} 
{ Std, Dev,}         & { 27.78}           & { 24.21}                & { 27.63}              & { 16.28}               & { 26.21}                & { \textbf{15.86}}       & { 17.15}                 & { 20.97}            & { 23.97}                & { 17.9}              & { \textbf{11.74}}       & { 12.55}       
\\ \bottomrule
\end{tabular}%
}
\caption{Activity-wise MPJPE scores for comparable works (with and without Procrustes transformation), including the top-3 in CVPR'17 Human 3.6 challenge and the top-3 IJCVm Jan'18. Columns marked with * make use of additional information or datasets, among the ones depicted in Table \ref{tab:recent}. Results in \textbf{bold} show the best MPJPE score among methods not relying on multiple datasets or additional information at training time. {\ul\textit{Underlined}} results show the best MPJPE score among all the methods, including the ones employing additional training time information. }
\label{tab:results}
\end{table*}

\begin{table}[]
\centering
\resizebox{.3\textwidth}{!}{%
\begin{tabular}{@{}lccccc@{}}
\toprule
\multicolumn{1}{l}{} & \multicolumn{1}{l}{ \textbf{Year}} & \multicolumn{1}{l}{\textbf{L.}} & \multicolumn{1}{l}{\textbf{T.}} & \multicolumn{1}{l}{\textbf{M.D.}} & \multicolumn{1}{l}{\textbf{D.A.}} \\ \toprule
\rowcolor[HTML]{E2EFDA} 
Cheng \cite{cheng20203d}                                        & 2020                                                                             &                                                                                & X                                                                              & X                                                                                & X                                                                                \\
Pham \cite{pham2020unified}                                         & 2019                                                                             & X                                                                              & X                                                                              & X                                                                                &                                                                                  \\
\rowcolor[HTML]{E2EFDA} 
Zhao \cite{zhao2019semantic}                                        & 2019                                                                             & X                                                                              &                                                                                & X                                                                                &                                                                                  \\
Chen \cite{chen2020anatomy}                                        & 2020                                                                             &                                                                                & X                                                                              &                                                                                  & X                                                                                \\
\rowcolor[HTML]{E2EFDA} 
Lin \cite{lin2019trajectory}                                          & 2019                                                                             & X                                                                              & X                                                                              &                                                                                  &                                                                                  \\
Sharma \cite{sharma2019monocular}                                       & 2019                                                                             & X                                                                              &                                                                                & X                                                                                & X                                                                                \\
\rowcolor[HTML]{E2EFDA} 
Tripathi \cite{tripathi2020posenet3d}                                     & 2020                                                                             & X                                                                              & X                                                                              &                                                                                  & X                                                                                \\
Wandt \cite{wandt2019repnet}                                        & 2019                                                                             & X                                                                              &                                                                                &                                                                                  & X                                                                                \\
\rowcolor[HTML]{E2EFDA} 
Arnab \cite{arnab2019exploiting}                                        & 2019                                                                             & X                                                                              & X                                                                              & X                                                                                &                                                                                  \\
Mehta \cite{mehta2020xnect}                                        & 2019                                                                             & X                                                                              & X                                                                              & X                                                                                &                                                                                  \\
\rowcolor[HTML]{E2EFDA} 
\textbf{DECA-H4}                                & 2020                                                                             &                                                                                &                                                                                &                                                                                  &                                                        
\\ \bottomrule
\end{tabular}%
}
\vspace*{7pt}
\caption{Comparison of the most relevant competing methods from 2019-2020 (top Average MPJPE on Human3.6M). \textbf{L.}: using 2D joints ground truth and/or lifting from 2D joints, \textbf{T.}: using temporal information, \textbf{M.D.}: using multiple training datasets, \textbf{D.A.}: using non-standard data augmentation techniques or biometric models. In the table we did not include works with lower Average MPJPE than ours.}
\label{tab:recent}
\end{table}

\subsection{Qualitative results}
\begin{figure*}[!ht]
\centering
\begin{minipage}[c]{.1\textwidth}
  \vspace*{\fill}
  \centering
  \includegraphics[height=1cm]{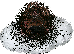}
  \subcaption{}
  \label{fig:qual_irt}\par\vfill
  \includegraphics[height=3cm]{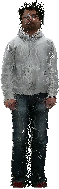}
  \subcaption{}
  \label{fig:qual_irf}
\end{minipage}%
\begin{minipage}[c]{.4\textwidth}
  \vspace*{\fill}
  \centering
  \includegraphics[height=4cm]{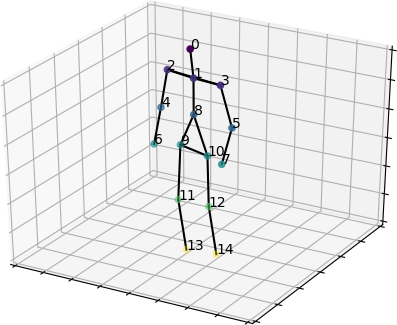}
  \subcaption{GT}
  \label{fig:qual_gt}
\end{minipage}%
\begin{minipage}[c]{.2\textwidth}
  \vspace*{\fill}
  \centering
  \includegraphics[height=2cm]{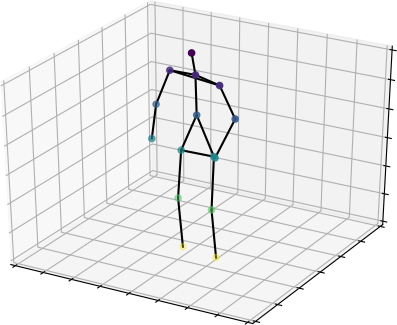}
  \subcaption{\{T\};\{T\}}
  \label{fig:qual_pt}\par\vfill
  \includegraphics[height=2cm]{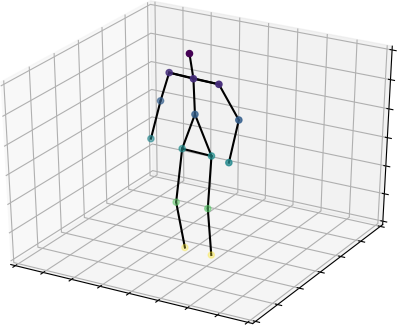}
  \subcaption{\{F\};\{F\}}
  \label{fig:qual_pf}
\end{minipage}%
\begin{minipage}[c]{.2\textwidth}
  \vspace*{\fill}
  \centering
  \includegraphics[height=2cm]{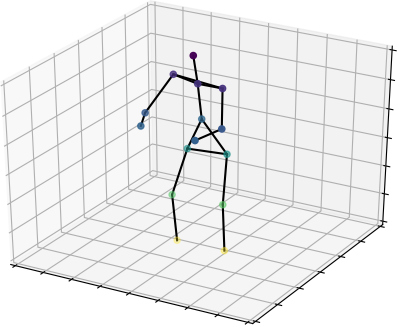}
  \subcaption{\{T\};\{F\}}
  \label{fig:qual_ptvt}\par\vfill
  \includegraphics[width=2cm,height=2cm]{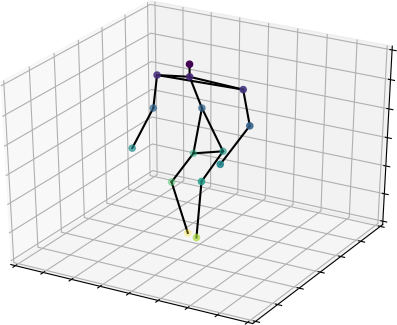}
  \subcaption{\{F\};\{T\}}
  \label{fig:qual_pfvt}
\end{minipage}
\caption{DECA-R4 qualitative results on the PanopTOP31K dataset. On the left (
\subref{fig:qual_irt}, 
\subref{fig:qual_irf}) the types of input accepted by DECA (top-view or front-view). DECA can also accept inputs in the depth domain. In the center (\subref{fig:qual_gt}), the corresponding 3D ground truth. On the right, the possible combinations of training/testing experiments. \textbf{T} stands for \textbf{top} and \textbf{F} stands for \textbf{front}. As an example, in (\subref{fig:qual_ptvt}), \textit{\{T\};\{F\}} means that DECA has been trained exclusively on \textbf{top} data and tested on previously unseen (not even at validation time) \textbf{front} data.}
\label{fig:qualitative}
\end{figure*}

In Fig. \ref{fig:qualitative} we show some qualitative results from DECA-R4 configuration on RGB data. We deploy our network training and testing on all the possible viewpoint combinations. The network takes as input either the top-view RGB (Fig. \ref{fig:qual_irt}) image or the front view (Fig. \ref{fig:qual_irf}) one. When trained and tested on the same viewpoint (Fig. \ref{fig:qual_pt}, \ref{fig:qual_pf}), the network produces similar outputs, thus confirming its ability to deal with the challenging top-view scenario. When training on the top view and testing on the front one (Fig. \ref{fig:qual_ptvt}), the network can accurately retrieve the positions of the lower body joints. DECA can retrieve parts of the body mostly occluded ad training time, thus displaying its generalization capabilities. When training on the front view and testing on the top one (Fig. \ref{fig:qual_pfvt}), the network can retrieve the positions of the upper body joints, which are visible in both images but from different perspectives, proving that DECA can internally model the viewpoint.

\begin{figure*}
\centering
\begin{minipage}[c]{.36\textwidth}
      \includegraphics[width=\textwidth]{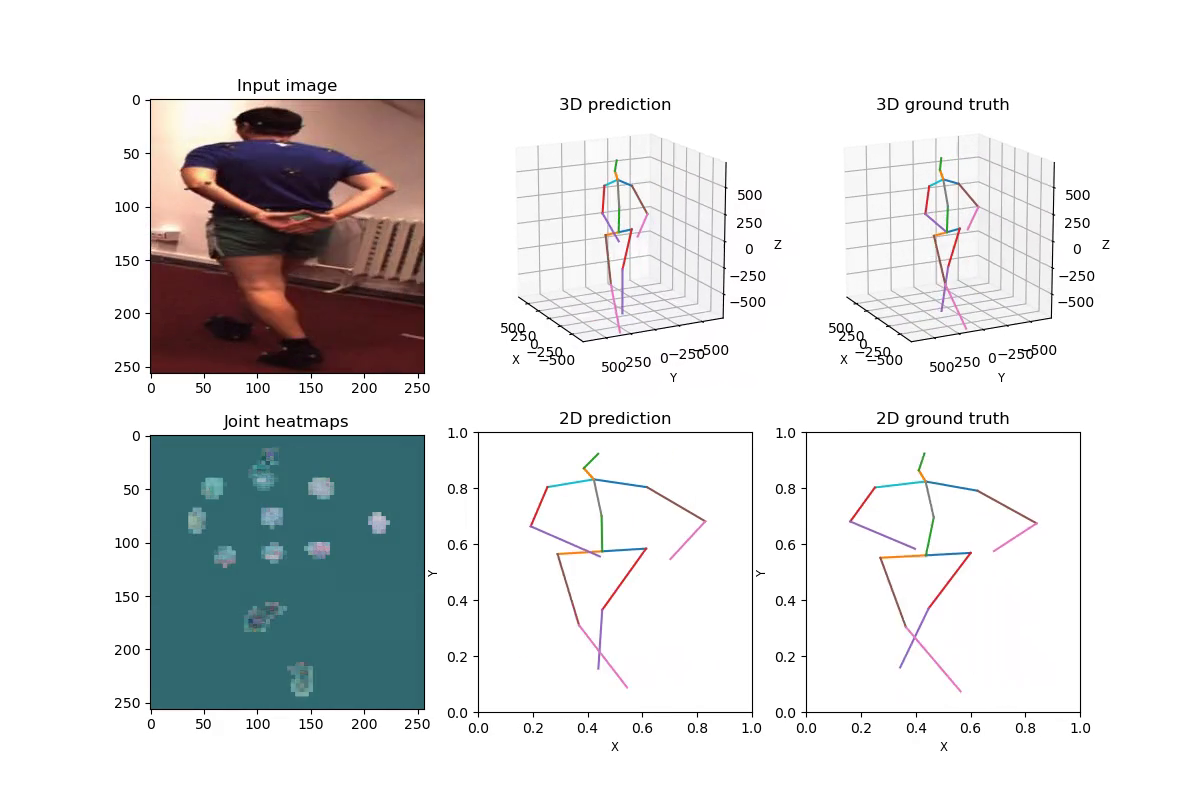}
      \subcaption{Results from 'Walking' activity.}
      \label{fig:walking}
\end{minipage}
\begin{minipage}[c]{.36\textwidth}
      \includegraphics[width=\textwidth]{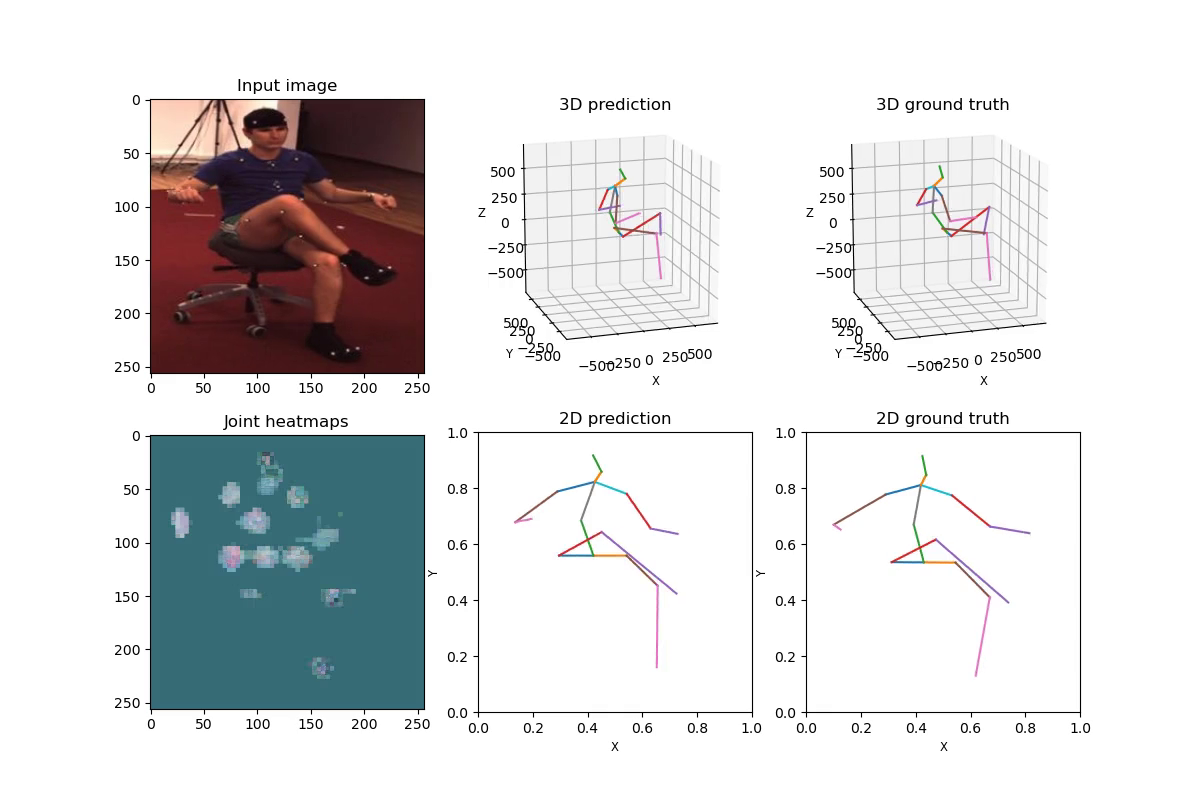}
      \subcaption{Results from 'Sitting Down' activity.}
        \label{fig:sittingdown}
\end{minipage}
\begin{minipage}[c]{.25\textwidth}
      \includegraphics[width=\textwidth]{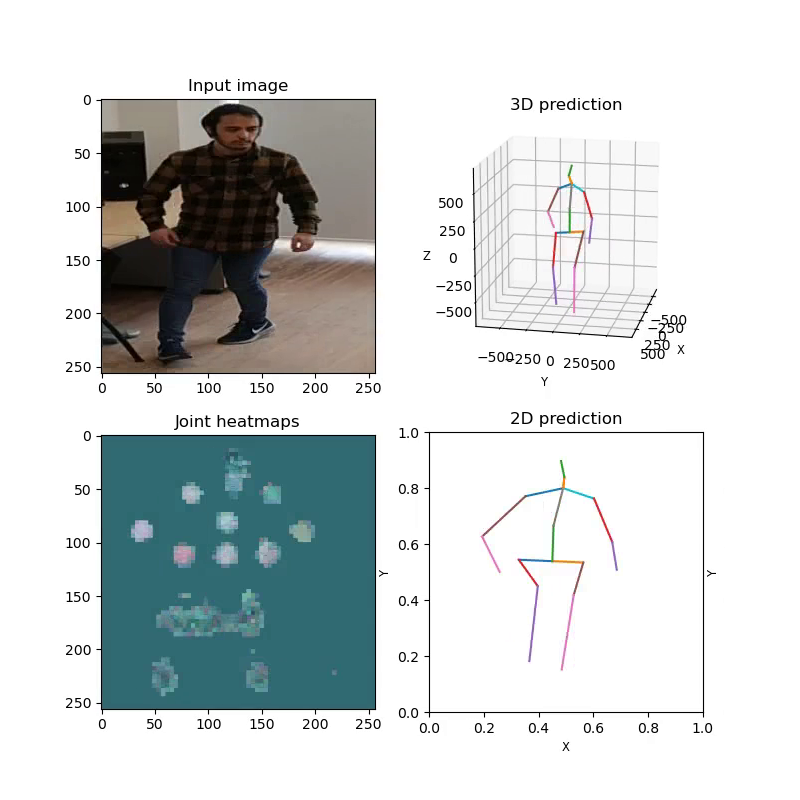}
      \subcaption{In-the-wild results.}
      \label{fig:wild}
\end{minipage}
    \caption{Qualitative results of DECA-H4 on the Human3.6M dataset (\subref{fig:walking}, \subref{fig:sittingdown}) and in-the-wild (\subref{fig:wild})}.
    \label{fig:qualitative_human}
\end{figure*}

In Figs. \ref{fig:walking}, \ref{fig:sittingdown} we show some qualitative results for the \textit{Walking} and \textit{Sitting Down} activities from test examples of the Human3.6 dataset. Starting from the upper left: input RGB image, predicted 3D pose, ground truth 3D pose, combination of the 17 "attention" heatmaps, predicted 2D pose and ground truth 2D pose. In Fig. \ref{fig:wild} we show some in-the-wild results (no ground truth is present in this case).

\section{Conclusions}
\label{sec:conclusions}

We presented DECA, a deep viewpoint-equivariant method for human pose estimation on single RGB/depth images using capsule autoencoders. We show how CapsNets are better suited to deal with the 3D nature of raw data and how they allow taking a step forward to viewpoint equivariance. We have shown how our method can effectively generalize and achieve state-of-the-art results in both RGB and depth domains, as well as in the viewpoint transfer task. 

\bibliographystyle{unsrt}  
\bibliography{references}

\end{document}